\documentclass{article}

\usepackage{microtype}
\usepackage{graphicx}
\usepackage{subfigure}
\usepackage{booktabs}

\usepackage{hyperref}

\usepackage[accepted]{utils/icml2025}

\usepackage{amsmath}
\usepackage{amssymb}
\usepackage{mathtools}
\usepackage{amsthm}

\usepackage[capitalize,noabbrev]{cleveref}

\theoremstyle{plain}

\theoremstyle{definition}

\theoremstyle{remark}

\usepackage[textsize=tiny]{todonotes}

\usepackage{utils/utils}

\usepackage{booktabs}
\usepackage{multirow}
\usepackage{adjustbox}
\usepackage{makecell}

\icmltitlerunning{}

\begin{document}

\twocolumn[
\icmltitle{Large Language-Geometry Model: When LLM meets Equivariance}

\begin{icmlauthorlist}
\icmlauthor{Zongzhao Li}{ruc}
\icmlauthor{Jiacheng Cen}{ruc}
\icmlauthor{Bing Su}{ruc}
\icmlauthor{Wenbing Huang}{ruc}
\icmlauthor{Tingyang Xu}{damo,hupan}
\icmlauthor{Yu Rong}{damo,hupan}
\icmlauthor{Deli Zhao}{damo,hupan}

\end{icmlauthorlist}

\icmlaffiliation{ruc}{Gaoling School of Artificial Intelligence, Renmin University of China}
\icmlaffiliation{damo}{DAMO Academy, Alibaba Group, Hangzhou, China}
\icmlaffiliation{hupan}{Hupan Lab, Hangzhou, China}

\icmlcorrespondingauthor{Wenbing Huang}{hwenbing@126.com}
\icmlcorrespondingauthor{Yu Rong}{yu.rong@hotmail.com}


\vskip 0.3in
]

\makeatletter
\renewcommand{\ICML@appearing}{}
\makeatother
\printAffiliationsAndNotice{}  %

\begin{abstract} 
Accurately predicting 3D structures and dynamics of physical systems is crucial in scientific applications.
Existing approaches that rely on geometric Graph Neural Networks (GNNs) effectively enforce $\mathrm{E}(3)$-equivariance, but they often fall in leveraging extensive broader information. While direct application of Large Language Models (LLMs) can incorporate external knowledge, they lack the capability for spatial reasoning with guaranteed equivariance. 
In this paper, we propose EquiLLM, a novel framework for representing 3D physical systems that seamlessly
integrates E(3)-equivariance with LLM capabilities.
Specifically, EquiLLM comprises four key components: geometry-aware prompting, an equivariant encoder, an LLM, and an equivariant adaptor. Essentially, the LLM guided by the instructive prompt serves as a sophisticated invariant feature processor, while 3D directional information is exclusively handled by the equivariant encoder and adaptor modules.
Experimental results demonstrate that EquiLLM delivers significant improvements over previous methods across molecular dynamics simulation, human motion simulation, and antibody design, highlighting its promising generalizability.
\end{abstract}

\section{Introduction}

Accurately predicting 3D structures/dynamics of physical systems remains a fundamental challenge in physics and biology. Typical tasks such as molecular dynamics simulation~\citep{hollingsworth2018molecular} and antibody design~\citep{tiller2015advances} require not only a deep understanding of complex spatial geometry but also the preservation of $\mathrm{E}(3)$-equivariance --- ensuring predictions transform correspondingly with input rotations, reflections and translations~\citep{batzner20223, huang2022equivariant}. From the machine learning perspective, $\mathrm{E}(3)$-equivariant models are more powerful than their non-equivariant counterparts, as they are inherently generalizable across arbitrary coordinate systems when modeling physical systems. To achieve equivariance, current approaches primarily rely on geometric Graph Neural Networks (GNNs)~\citep{wu2024equivariant, kong2022conditional}. Despite their fruitful progress, these models often lack the ability to leverage external domain knowledge and broader contextual information, such as task-specific instructions and expert-curated guidance, hindering further performance enhancement.

Recently, Large Language Models (LLMs) have demonstrated remarkable success across a wide range of applications, owing to their large-scale pretraining on extensive datasets and their substantial model size. It is well known that LLMs can not only understand and generate text but also excel at integrating and leveraging scientific knowledge~\citep{liu2025integrating, jablonka2024leveraging, wang2023scibench}. For instance, LLMs can comprehend fundamental chemical concepts and molecular structural characteristics~\citep{guo2023can}. More significantly, LLMs' flexibility in prompt engineering enables the development of tailored instructions that better leverage their capabilities, producing outputs more precisely suited to the task.

A natural idea is to directly employ LLMs for modeling 3D physical systems. However, this approach \emph{fails to} yield satisfactory results in practice. A key limitation is that LLMs are trained to process ordered and discrete text tokens, restricting their ability to directly comprehend unordered and continuous data in 3D space. One possible solution is to adapt existing multimodal LLM architectures, such as LLaVA~\citep{liu2024visual}, by treating 3D structures as a separate modality and simply replacing the image encoder with a geometric GNN. However, this naive adaptation fails to satisfy the $\mathrm{E}(3)$-equivariance requirement. Since geometric GNNs produce both invariant features and equivariant coordinates, passing these outputs through an LLM inevitably compromises equivariance. \emph{Therefore, it is non-trivial to integrate the strengths of both LLMs and geometric GNNs while maintaining essential geometric properties}.

\begin{figure*}
    \centering
    \includegraphics[width=0.98\textwidth]{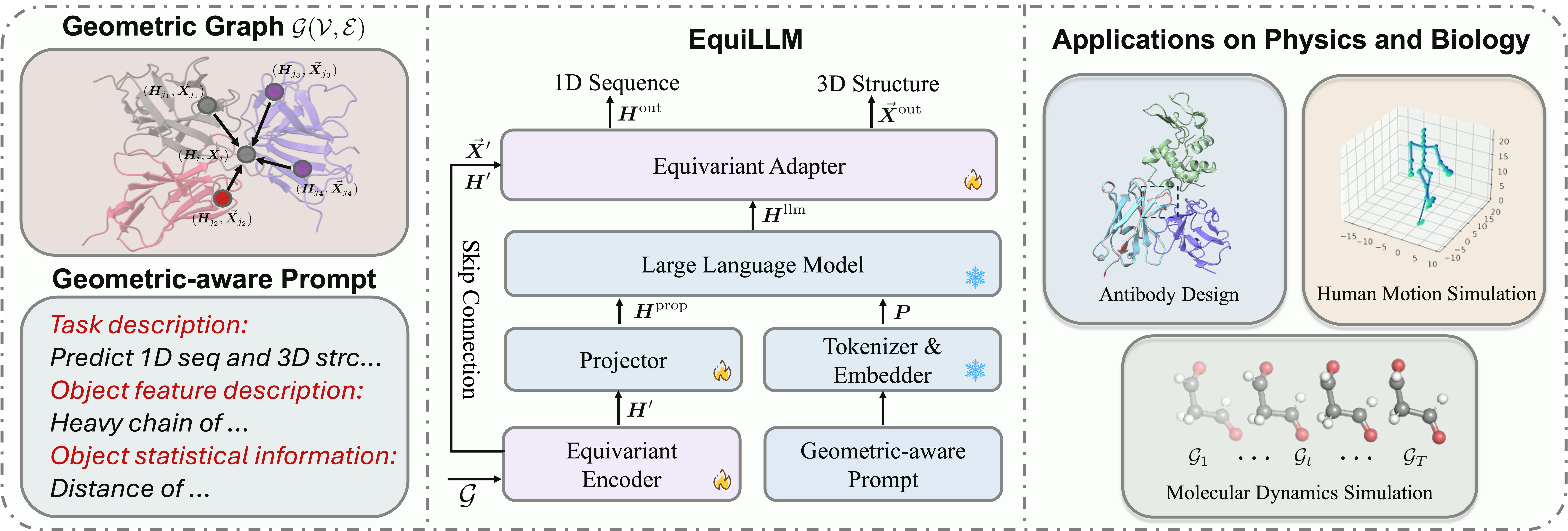}
    \caption{The overall framework of EquiLLM. Given a geometric graph $\gG = (\gV,\gE)$ as input, EquiLLM initially employs an Equivariant Encoder to derive processed features $\Vec{\mX}^{'}$ and $\mH^{'}$. The features $\mH^{'}$ are first projected through a projector, then concatenated with prompt features $\mP$ in a task-specific manner. This concatenated vector is subsequently fed into an LLM. The output features $\mH^{\text{llm}}$ from the LLM, alongside the previously obtained processed features $\Vec{\mX}^{'}$ and $\mH^{'}$, are then passed into an Equivariant Adapter. The Equivariant Adapter then generates the final outputs, including the vector $\Vec{\mX}^{\text{out}}$ for equivariant tasks and the feature $\mH^{\text{out}}$ for invariant tasks. The blue \colorbox{9CBCE3!30}{\texttt{module}} means the invariant module, while the purple \colorbox{8A2BE2!10}{\texttt{module}} means the equivariant module.}
    \label{fig:architecture}
\end{figure*}

To this end, this paper introduces EquiLLM, a novel framework for representing 3D physical systems that seamlessly integrates $\mathrm{E}(3)$-equivariance with LLM capabilities. 
EquiLLM is carefully designed and comprises four core modules (see Figure \ref{fig:architecture}): geometry-aware prompting, an equivariant encoder, an LLM, and an equivariant adapter. 
A key insight of EquiLLM in maintaining equivariance lies in its innovative design: the LLM guided by the instructive prompt serves as a sophisticated invariant feature processor, while 3D directional information is exclusively handled by the equivariant encoder and adaptor modules.
Specifically, to fully activate the spatial reasoning capabilities of LLMs, we introduce the geometry-aware prompts containing invariant geometric information, including task description, input
feature description and statistical information. Then, EquiLLM employs a geometric GNN as a domain-specific encoder to effectively model and extract 3D representations of input systems. Subsequently, to satisfy the equivariance constraint, the input to the LLM are strictly limited to the prompt and invariant features derived from the equivariant encoder's output. The LLM-generated outputs are subsequently combined with the 3D equivariant vectors produced by the equivariant encoder and fed into an equivariant adaptor for information fusion. EquiLLM ultimately produces both invariant labels and equivariant coordinates required by downstream applications.

To sum up, our main contributions are threefold:
\begin{itemize}
    \item To the best of our knowledge, we present the first investigation into modeling 3D physical systems by integrating LLMs with geometric GNNs, aiming to combine the strengths of both approaches.
    \item We present EquiLLM, a novel framework that is meticulously designed to permit E(3)-equivariance and instill 3D spatial reasoning into LLMs' powerful capabilities. 
    \item We conduct extensive experiments on diverse tasks of molecular dynamics simulation, human motion simulation and antibody design. The results show that our method achieves superior performance, attaining state-of-the-art results in nearly all metrics. 
\end{itemize}

\section{Related Work}
\textbf{Geometric GNN.} Due to its thorough consideration of physical symmetries in 3D space, Geometric GNN~\citep{satorras2021n, fuchs2020se} has been widely applied in various scientific tasks, such as dynamics simulation~\citep{xu2024equivariant} in physics and antibody design~\citep{lin2024geoab} in biology. ESTAG~\citep{wu2024equivariant} enhances the model's simulation capability for object dynamics trajectories in both spatial and temporal dimensions through interleaved Equivariant Spatial Module and Equivariant Temporal Mod   ule. SEGNO~\citep{liu2024segno} incorporates the second-order graph neural ODE with equivariant property to reduce the roll-out error of long-term physical simulation. MEAN~\citep{kong2022conditional} not only leverages an Internal context encoder to model spatial correlations within antibody chains but also introduces an External attentive encoder with an attention mechanism to better capture inter-chain relationships. GeoAB~\citep{lin2024geoab} extends the equivariant model GMN~\citep{huang2022equivariant} to handle binary, ternary, and quaternary interactions and applies it in atom-level updates. Despite significant progress these methods have made, they fail to explore how to harness the capabilities of LLMs to augment model performance.

\textbf{LLM + GNN.} Large Language Models (LLMs) with rich knowledge are being widely transferred and applied across multiple domains to enhance model capabilities~\citep{singhal2023large, singhal2025toward}. Numerous excellent works have emerged in combining GNNs with LLMs for scientific applications. ChemLLMBench~\citep{guo2023can} tests LLM's understanding, reasoning, and explaining capabilities on various chemical tasks using in-context learning. Prot2Text~\citep{abdine2024prot2text} integrates protein sequence, structure, and textual annotations into an encoder-decoder framework composed of GNN and LLM to predict protein functions. MoleculeSTM~\citep{liu2023multi} uses a contrastive learning paradigm to align molecular graphs and textual descriptions in the semantic space, thereby learning better feature representations. MolCA~\citep{liu2023molca} employs Q-Former~\citep{li2023blip} as a cross-modal projector to align the feature spaces of graph encoder and language encoder, enhancing performance in molecule captioning tasks. Although the aforementioned methods promote interactions between GNNs and LLMs through various paradigms and yield promising results, they have yet to explore tasks involving 3D structural data, such as 3D structure generation and dynamic trajectory simulation in 3D space. The EquiLLM framework we propose in this paper is the first to integrate LLMs with Geometric GNNs that embed spatial symmetry constraints and has been effectively validated across datasets from both physical and biological domains.

\section{Method}
In this section, we first introduce the preliminaries related to geometric modeling in \cref{sec:3.1}. Next, in \cref{sec:3.2}, we present the proposed framework EquiLLM. Finally, in \cref{sec:3.3}, we describe how the EquiLLM framework is applied to two representative tasks (\emph{e.g.} dynamic simulation and antibody design). The overview of our EquiLLM is illustrated in \cref{fig:architecture}.

\subsection{Preliminaries, Notations and Definitions}
\label{sec:3.1}
Physical systems (such as molecules) can be naturally modeled with geometric graphs. We represent each static system as a geometric graph $\gG = (\gV,\gE)$, where each node $v_i$ in $\gV$ is associated with an invariant feature $\vh_{i}\in\R^c$ (\emph{e.g.} atom type) and an 3D equivariant vector $\vec{\vx}_{i}\in\R^3$ (\emph{e.g.} atom coordinates); each edge (\emph{e.g.} chemical bonds) denotes the connectivity between nodes. Apart from modeling static systems, we explore dynamic systems, focusing on constructing geometric graphs across different time steps. The details will be thoroughly discussed in \cref{sec:3.3.1}. In the following sections, we use the matrices $\Vec{\mX}\in\R^{N \times 3}$ and $\mH\in\R^{N \times c}$ to denote the sets of node coordinates and invariant features of the geometric graph $\gG$.

\textbf{Task Formulation.} Here, we provide a general form of our task and will elaborate specific applications including dynamic simulation and antibody design in \cref{sec:3.3}. Given the input geometric graph $\gG^{\text{in}}$, our goal is to find a function $\phi$ to predict the output $\gG^{\text{out}}$. This process can be formally delineated as:
\begin{equation}\label{eq:task_general}
\begin{aligned}
    &\gG^{\text{out}} = \phi(\gG^{\text{in}}). \\
\end{aligned}
\end{equation}
Meanwhile, since we introduce LLMs into our framework, we will further construct task-specific prompts to guide the extraction of relevant domain knowledge from LLMs, recasting our task as:
\begin{equation}\label{eq:dynamic2}
\begin{aligned}
    &\gG^{\text{out}} = \phi(\gG^{\text{in}}, \mP), \\
\end{aligned}
\end{equation}
where $\mP$ denotes the prompt.

\textbf{Equivariance.} It is crucial to emphasize that in the tasks above, the function $\phi$ must satisfy $\mathrm{E}(3)$ symmetries of physical laws. Specifically, if arbitrary translations, reflections, or rotations are applied to the input coordinate matrix $\Vec{\mX}^{\text{in}}$, the output coordinate matrix $\Vec{\mX}^{\text{out}}$ should undergo the corresponding transformation.

\subsection{Large Language Geometric Model}
\label{sec:3.2}
In this section, we provide a meticulous description of our model EquiLLM, which consists of three main components: Equivariant Encoder, LLM, and Equivariant Adapter. Unlike existing works~\citep{gruver2024finetuned} that applys an LLM to predict the 3D coordinates directly, EquiLLM leverages an LLM to acquire broader scientific domain knowledge while employing geometric GNNs for precise modeling of 3D structures. These two components are seamlessly integrated through an equivariant adapter, achieving superior predictive performance without compromising $\mathrm{E}(3)$-equivariance.

\textbf{Equivariant Encoder.} The Equivariant Encoder is a domain-specific equivariant model, which can be any suitable equivariant model from the relevant field. The model takes the graph $\gG^{\text{in}} = (\gV^{\text{in}},\gE^{\text{in}})$ as input, performing initial encoding and embedding of geometric information, and outputs a processed geometric graph $\gG' = (\gV',\gE')$, This process can be formally defined as:
\begin{equation}\label{eq:equ_encoder}
\begin{aligned}
    &\gG' = \phi_{e} (\gG^{\text{in}}), \\
\end{aligned}
\end{equation}
where $\phi_{e}$ can be any equivariant model, used to jointly model the geometric relationships between $\Vec{\mX}^{\text{in}}$ and $\mH^{\text{in}}$ features across different nodes, resulting in processed features $\Vec{\mX}^{'}$ and $\mH^{'}$.

Since LLMs are not naturally equivariant, directly feeding $\mX'$ into an LLM would likely undermine the intrinsic equivariance of the overall architecture. Thus, in contrast to existing works, we convey the invariant features $\mH'$ to the LLM, but pass the equivariant matrix $\mX'$ to the subsequent Equivariant Adapter via a skip connection. Before feeding $\mH'$ to the LLM, we first conduct a projector on $\mH'$ to align its dimension with the input space of the LLM. This process can be formally characterized as:
\begin{equation}\label{eq:equ_encoder_proj}
\begin{aligned}
    \mH^{\text{proj}} = \phi_{\text{proj}} (\mH'), \\
\end{aligned}
\end{equation}
where $\phi_{\text{proj}}$ is implemented as a linear layer in EquiLLM.

\textbf{Geometric-aware Prompt.} One may directly input the aligned features $\mH^{\text{proj}}$ into the LLM to make the final predictions. However, this approach overlooks the pivotal role of the prompt, as it does not utilize the linguistic form of the prompt to effectively harness the LLM's comprehension and articulation of the specific task at hand. Therefore, in the EquiLLM framework, we carefully design task-specific prompts for different tasks to unleash domain-specific knowledge. 

The prompt content for all tasks can be broadly divided into three key components: (1) \emph{task description}, (2) \emph{object feature description}, and (3) \emph{object statistical information}.

\textbf{$\triangleright$ Task description.} The task description consists of two parts: <Task> and <Requirement>. <Task> appears at the beginning of the prompt, providing a succinct description of the task to help the LLM quickly identify the task's objective. <Requirement> is located in the main body of the prompt and elaborates on the input-output requirements and constraints of the task, ensuring a comprehensive understanding of the task by the LLM. 

\textbf{$\triangleright$ Object feature description.} The feature description of the input object begins with <Object> and primarily outlines the composition information as well as the structural characteristics of the input object. 

\textbf{$\triangleright$ Object statistical information.} This components starts with <Statistics>, encapsulating detailed metrics pertaining to the distribution of the object's coordinates in 3D space, including the maximum, minimum, and mean values. It is crucial to note that, unlike conventional tasks, directly incorporating absolute coordinate values into the prompt is not recommended in 3D spatial modeling tasks. This is due to the fact that transformations such as translation, reflection, or rotation applied to the input object will invariably alter the corresponding coordinate distribution, thereby violating the principle that the prompting process must remain $\mathrm{E}(3)$-invariant. Consequently, we represent the coordinate distribution of the input object indirectly by computing statistical metrics related to distances.

\textbf{Large Language Model (LLM).} After designing the prompt, we employ the tokenizer and embedding layer of the LLM to obtain the corresponding word embedding features, denoted as $\mP$. Subsequently, depending on the specific task, we concatenate $\mP$ with the invariant features $\mH^{\text{proj}}$ in an appropriate way. The concatenation strategies for different tasks will be discussed in detail in \cref{sec:3.3}. Next, the concatenated features are input into the LLM with the aim of unlocking and leveraging the scientific knowledge embedded within the LLM to enhance the model's understanding and reasoning capabilities for the relevant tasks. Unlike previous works like CrystalLLM~\citep{gruver2024finetuned} and UniST~\citep{yuan2024unist} that require fine-tuning some layers within the LLM, resulting in significant computational costs and time expenditure, the EquiLLM framework freezes all parameters of the LLM, eliminating the need for additional training. This process can be roughly represented as follows:
\begin{equation}\label{eq:llm}
\begin{aligned}
    &[\mH^{\text{llm}}, \mP^{\text{llm}}] = \mathbf{LLM} (\texttt{Concat} (\mH^{\text{proj}}, \mP)), \\
\end{aligned}
\end{equation}

\textbf{Equivariant Adapter.} Upon obtaining the output from the LLM, we extract the part corresponding to the invariant features, denoted as $\mH^{\text{llm}}$. While directly utilizing it for final predictions may be viable for invariant tasks (\emph{e.g.} predicting the energy of a molecular system), it is inadequate for equivariant tasks, where the core objective is to predict the 3D coordinates of objects. To address this challenge, we propose the Equivariant Adapter, which leverages one-layer EGNN to process $\mH^{\text{llm}}$ while minimizing the introduction of excessive additional parameters. Specifically, we first employ a projection layer to re-project $\mH^{\text{llm}}$ back into the space corresponding to the invariant features $\mH'$ and add it with $\mH'$, yielding the refined feature representation $\mH^{r}$. Then, both the equivariant coordinate features $\Vec{\mX'}$ from Equivariant Encoder and the refined invariant features $\mH^{r}$ are transmitted to the EGNN, yielding the output $\Vec{\mX}^{\text{out}}$ and $\mH^{\text{out}}$. The whole process is formally expressed as:
\begin{equation}\label{eq:EA}
\begin{aligned}
    &\vm_{ij} = \varphi_m \left(\vh_{i}^{r},\, \vh_{j}^{r},\, \left\| \vec{\vx}'_{i} - \vec{\vx}'_{j} \right\|\right), \\
    &\vh_{i}^{\text{out}}  = \textstyle\vh_{i}^{r} + \varphi_{h}\left(\vh_{i}^{r},\sum_{j \in \mathcal{N}(i)}\vm_{ij}\right), \\
    &\vec{\vx}_{i}^{\text{out}} = \textstyle\vec{\vx}'_{i}+\frac{1}{|\mathcal{N}(i)|}\sum_{j \in \mathcal{N}(i)}\varphi_{x} \left(\vm_{ij}\right)\cdot\left(\vec{\vx}'_{i}-\vec{\vx}'_{j}\right), \\ 
\end{aligned}
\end{equation}
where $\varphi_{m}$, $\varphi_{x}$, and $\varphi_{h}$ denote Multi-Layer Perceptrons (MLPs), and $\mathcal{N}(i)$ refers to the set of neighboring nodes associated with the $i$-th node. Specifically, $\vm_{ij}$ represents an $\mathrm{E}(3)$-invariant message transmitted from node $j$ to node $i$, which is utilized to aggregate and refine the feature vector $\vh_{i}^{r}$ via the function $\varphi_{h}$. Regarding the update of $\vec{\vx}'_{i}$, the function $\varphi_{x}$ is employed to compute a scalar $\varphi_{x} (\vm_{ij})$, which is subsequently multiplied by the difference $\vec{\vx}'_{i} - \vec{\vx}'_{j}$ to retain directional information, while incorporating residual connections to ensure translation equivariance.

Our EquiLLM framework guarantees that the overall architecture preserves the critical property of $\mathrm{E}(3)$-equivariance in 3D space while also avoiding the introduction of lengthy text context due to direct 3D coordinate input, which could severely affect the efficiency of training and inference. Moreover, compared with domain-specific Equivariant Encoder, EquiLLM introduces only two projection layers and a one-layer EGNN network, significantly reducing the additional training parameters in comparison to the existing literature~\citep{jin2024timellm, yuan2024unist}. Finally, our EquiLLM framework demonstrates exceptional flexibility and can be applied to various geometric modeling tasks, showcasing its robustness and generalizability.

\subsection{Applications on Dynamic Simulation and Antibody Design}
\label{sec:3.3}
In this section, we will present a detailed discussion on the application of our EquiLLM in dynamic simulation and antibody design.
\subsubsection{Dynamic Simulation}
\label{sec:3.3.1}
While our model is applicable to the simulations of both molecular dynamics and human motions, we illustrate molecular dynamics here as an example. Given the 3D coordinate trajectory of a physical system (\emph{e.g.,} molecules) over $T$ frames, namely, $\Vec{\mX}\in\mathbb{R}^{T \times N \times 3}$, along with the invariant features $\mH\in\mathbb{R}^{N \times c_{a}}$ encoded by atomic numbers, the model aims to infer future trajectories $\Vec{\mX}\in\mathbb{R}^{F \times N \times 3}$ for $F$ subsequent frames.

\textbf{Geometric-aware Prompt.}
Here, we will provide a general overview of the contents encompassed within the prompt.

\textbf{$\triangleright$ Task description.} The model is tasked with predicting the 3D coordinates $(x, y, z)$ of heavy atoms for the next $F$ frames based on the information from the previous $T$ frames.

\textbf{$\triangleright$ Object feature description.} For molecular systems, the emphasis is on compositional information and structural characteristics.

\textbf{Task Formulation and Training Objective. }
With $\gG_{1:T}\coloneqq\{\gG_{t} = (\Vec{\mX}_{t}, \mH_{t}, \gE)\}_{t=1}^{T}$ and $\gG_{T+1:T+F}\coloneqq\{\gG_{t} = (\Vec{\mX}_{t}, \mH_{t}, \gE)\}_{t=T+1}^{T+F}$, we provide the entire process as follows:
\begin{equation}\label{eq:task_dynamic}
\begin{aligned}
    \gG_{T+1:T+F} = \phi (\gG_{1:T}, \mP). \\
\end{aligned}
\end{equation}
Let $\Vec{\mX}^{\text{gt}}_{T+f}$ denote the ground-truth 3D coordinates for the time period from $T+1$ to $T+F$, we define the object function as $\textstyle\mathcal{L} = \frac{1}{|F|}\sum^{F}_{f=1}\ell_{\mathrm{mse}}(\Vec{\mX}^{\text{out}}_{T+f}, \Vec{\mX}^{\text{gt}}_{T+f})$ refers to the mean squared error (MSE).

\subsubsection{Antibody Design}
\label{sec:3.3.2}
Antibodies are Y-shaped proteins primarily responsible for recognizing and binding to specific antigens. Current research predominantly focuses on the variable region. The variable region is present in both the heavy and light chains of the antibody and can be further subdivided into the framework region and three Complementarity-Determining Regions (CDRs). These six CDRs are critical in determining the affinity between the antibody and antigen, with the CDR-H3 region on the heavy chain exhibiting the most pronounced variability. Consequently, the primary objective of this paper is to predict the amino acid sequence and the 3D coordinates of the CDR-H3 region, given the antibody-antigen complexes excluding the CDR-H3 region. In antibody design task, each node in $\gV$ associates with a trainable feature $\vh_i\in\R^{c_r}$ encoded by amino acid type and a matrix of 3D coordinates $\Vec{\mZ}_i \in \R^{4 \times 3}$. We choose 4 backbone atoms $\{\text{N}, \text{C}_\alpha, \text{C}, \text{O}\}$ to constitute $\Vec{\mZ}_i$.

\textbf{Geometric-aware Prompt.}
Here, we will provide a general overview of the contents encompassed within the prompt.

\textbf{$\triangleright$ Task description.} The model is tasked with predicting both the 1D sequence and 3D coordinates of CDR-H3 region. 

\textbf{$\triangleright$ Object feature description.} The structural features of the light chain, heavy chain, and antigen, which are described individually.

\textbf{Task Formulation and Training Objective.}
With $\gG = (\Vec{\mX}, \mH, \gE)$, where $(\Vec{\mX}$, $\mH )\coloneqq \{ (\Vec{\mZ_{i}}, \vh_{i}) \} ^{N}_{i=1}$, the entire process is delineated as follows:
\begin{equation}\label{eq:task_antibody}
\begin{aligned}
    &\mH^{\text{out}}, \Vec{\mX}^{\text{out}} = \phi_{r} (\gG, \mP_{r}), \\
    &\vy^{\text{out}} = \texttt{Softmax}(\mH^{\text{out}}), \\
\end{aligned}
\end{equation}
where $N$ denotes the number of residues in CDR-H3 region; $\vy^{\text{out}}$ and $\vy^{\text{gt}}$ denote the predicted distribution over all amino acid categories and the ground truth amino acid type; $\Vec{\mX}^{\text{out}}$ and $\Vec{\mX}^{\text{gt}}$ denote the predicted 3D structure and the ground truth 3D structure of the CDR-H3 region, respectively. The loss function is definited as $\mathcal{L} = \mathcal{L}_{\mathrm{seq}} + \lambda \mathcal{L}_{\mathrm{struct}}$, where $\mathcal{L} _{\mathrm{ce}}= \frac{1}{N}\ell_{ce}(\vy^{\text{out}}, \vy^{\text{gt}})$ denotes the cross entropy and $\mathcal{L} _{\mathrm{huber}}\frac{1}{N}\ell_{\mathrm{huber}}(\Vec{\mX}^{\text{out}}, \Vec{\mX}^{\text{gt}})$ denotes the Huber loss~\citep{huber1992robust}; the $\lambda$ is used to balance the two losses.

\section{Experiments}
We validate the effectiveness of the proposed EquiLLM framework on two tasks from different domains: the dynamic simulation in physics (\cref{sec:4.1}) and the antibody design in biology (\cref{sec:4.2}). Furthermore, in \cref{sec:4.3}, we conduct ablation studies, explore potential design variations of the EquiLLM framework, and discuss why the current framework is superior.

\subsection{Dynamic Simulation}
\label{sec:4.1}
In the dynamic simulation task, to demonstrate the broad applicability of our model across varying scales, we conduct experiments on two distinct datasets: the molecular-level MD17~\citep{chmiela2017machine} dataset and the macro-level Human Motion Capture~\citep{de2009guide} dataset. In order to expedite the dynamics simulations, we implement a sampling strategy based on previous research~\citep{huang2022equivariant} to extract a subset of trajectories for the purposes of training, validation, and testing. This approach involves randomly selecting an initial point and then sampling $2 \times T$ timestamps. The first $T$ timestamps are utilized as input for the models, while the remaining $T$ timestamps represent the future states that the models need to predict.

\textbf{Baselines and metrics.}
We evaluate EquiLLM with several baseline models, including traditional GNNs: ST\_GNN~\citep{gilmer2017neural} and STGCN~\citep{yu2017spatio}, equivariant GNNs: ST\_TFN~\citep{thomas2018tensor}, ST\_SE(3)-Tr.~\citep{fuchs2020se}, ST\_EGNN~\citep{satorras2021n}, and ESTAG~\citep{wu2024equivariant}. We also compare EquiLLM with existing LLMs, including GPT-4o-mini~\citep{openai20244o}, Gemini-1.5-flash-latest~\citep{team2024gemini}, and DeepSeek-V3~\citep{liu2024deepseek}. The models marked with ``ST'' are those we modified to handle multi-frame inputs by adding basic spatio-temporal aggregation, as done in ~\citep{wu2024equivariant}. For evaluation, we calculate the Mean Squared Errors (MSEs) averaged across all predicted frames as the metric.

\subsubsection{Molecular Dynamics}
\begin{table*}[!t]
\centering
\caption{Predicted MSE ($\times 10^{-3}$) on MD17 dataset.}
\label{table:md17_r10}
\begin{small}
\tabcolsep=0.09cm
\resizebox{\textwidth}{!}{
\begin{tabular}{lcccccccc}
\toprule
 & Aspirin  & Benzene & Ethanol & Malonaldehyde & Naphthalene & Salicylic & Toluene & Uracil \\
\midrule
\addlinespace[0.3em]
ST\_GNN &7.180\tiny{$\pm$0.003}	&1.359\tiny{$\pm$0.001}	&2.108\tiny{$\pm$0.001}	&5.620\tiny{$\pm$0.018}	&2.397\tiny{$\pm$0.017} &2.646\tiny{$\pm$0.003}	&2.233\tiny{$\pm$0.011}	&1.913\tiny{$\pm$0.012} \\
\addlinespace[0.3em]
ST\_TFN &7.389\tiny{$\pm$0.139}	&1.849\tiny{$\pm$0.003}	&2.041\tiny{$\pm$0.001}	&5.346\tiny{$\pm$0.006}	&3.555\tiny{$\pm$0.110}	&5.728\tiny{$\pm$0.015}	&2.979\tiny{$\pm$0.167}	&4.272\tiny{$\pm$0.035} \\
\addlinespace[0.3em]
STGCN &21.08\tiny{$\pm$0.001}	&654.7\tiny{$\pm$0.001}	&7.102\tiny{$\pm$0.001}	&32.87\tiny{$\pm$0.001}	&5.421\tiny{$\pm$0.001}	&3.501\tiny{$\pm$0.001}	&3.679\tiny{$\pm$0.001}	&7.142\tiny{$\pm$0.001} \\
\addlinespace[0.3em]
ST\_SE(3)-Tr. &6.234\tiny{$\pm$0.019}	&1.835\tiny{$\pm$0.001}	&1.765\tiny{$\pm$0.001}	&5.277\tiny{$\pm$0.070}	&3.256\tiny{$\pm$0.018}	&4.737\tiny{$\pm$0.016}	&2.104\tiny{$\pm$0.011}	&3.900\tiny{$\pm$0.006} \\
\addlinespace[0.3em]
ST\_EGNN &6.682\tiny{$\pm$0.380}	&1.482\tiny{$\pm$0.161}	&2.145\tiny{$\pm$0.001}	&4.729\tiny{$\pm$0.029}	&4.034\tiny{$\pm$0.028}	&6.296\tiny{$\pm$0.157}	&2.881\tiny{$\pm$0.002}	&3.394\tiny{$\pm$0.267} \\
\midrule
\addlinespace[0.3em]
GPT-4o-mini &13.070	&9.581	&5.011	&9.910 &35.155 &10.627 &8.132	&9.762 \\
\addlinespace[0.3em]
Gemini-1.5-flash-latest &17.347	&7.586	&8.871	&15.495 &15.188 &17.978 &15.426	&16.935 \\
\addlinespace[0.3em]
DeepSeek-V3$^{\star}$ &12.009	& 3.657	&6.729	&10.247 &7.933 &10.415 &6.941	&12.956 \\
\midrule
\addlinespace[0.3em]
ESTAG &3.263\tiny{$\pm$0.065}	&0.891\tiny{$\pm$0.083}	&1.090\tiny{$\pm$0.001}	&2.046\tiny{$\pm$0.085} &2.036\tiny{$\pm$0.350} &3.134\tiny{$\pm$0.094}	&1.634\tiny{$\pm$0.149}	&1.852\tiny{$\pm$0.066} \\
\addlinespace[0.3em]
EquiLLM &\textbf{2.391}\tiny{$\pm$0.233}	&\textbf{0.732}\tiny{$\pm$0.058}	&\textbf{1.031}\tiny{$\pm$0.001}	&\textbf{1.671}\tiny{$\pm$0.025}
&\textbf{1.453}\tiny{$\pm$0.071}		&\textbf{2.162}\tiny{$\pm$0.137}	&\textbf{1.178}\tiny{$\pm$0.186}	&\textbf{1.060}\tiny{$\pm$0.194} \\
\addlinespace[0.3em]
Reduction \emph{w.r.t} ESTAG &\textcolor{red}{-26.72\%}&\textcolor{red}{-17.84\%}	&\textcolor{red}{-5.41\%}	&\textcolor{red}{-18.32\%} &\textcolor{red}{-28.63\%} &\textcolor{red}{-31.01\%}	&\textcolor{red}{-27.90\%}	&\textcolor{red}{-42.76\%} \\
\bottomrule
\end{tabular}
}
\end{small}
\vspace{-0.2cm}
\end{table*}

\textbf{Implementation details.}
MD17 consists of time-evolving paths produced through molecular dynamics simulation for eight different small compounds (such as aspirin, benzene, and others). To ensure a fair comparison, all hyperparameters (\emph{e.g.} learning rate, number of training epochs) are kept consistent across our model and all other baselines. We utilize the ESTAG~\citep{wu2024equivariant} as the Equivariant Encoder and GPT-2~\citep{radford2019language}~\footnote{A more powerful LLM may lead to a superior performance. Here we use GPT-2 for concept validation.} as the language model within our EquiLLM framework.

\begin{figure}
  \centering
  \includegraphics[width=\columnwidth]{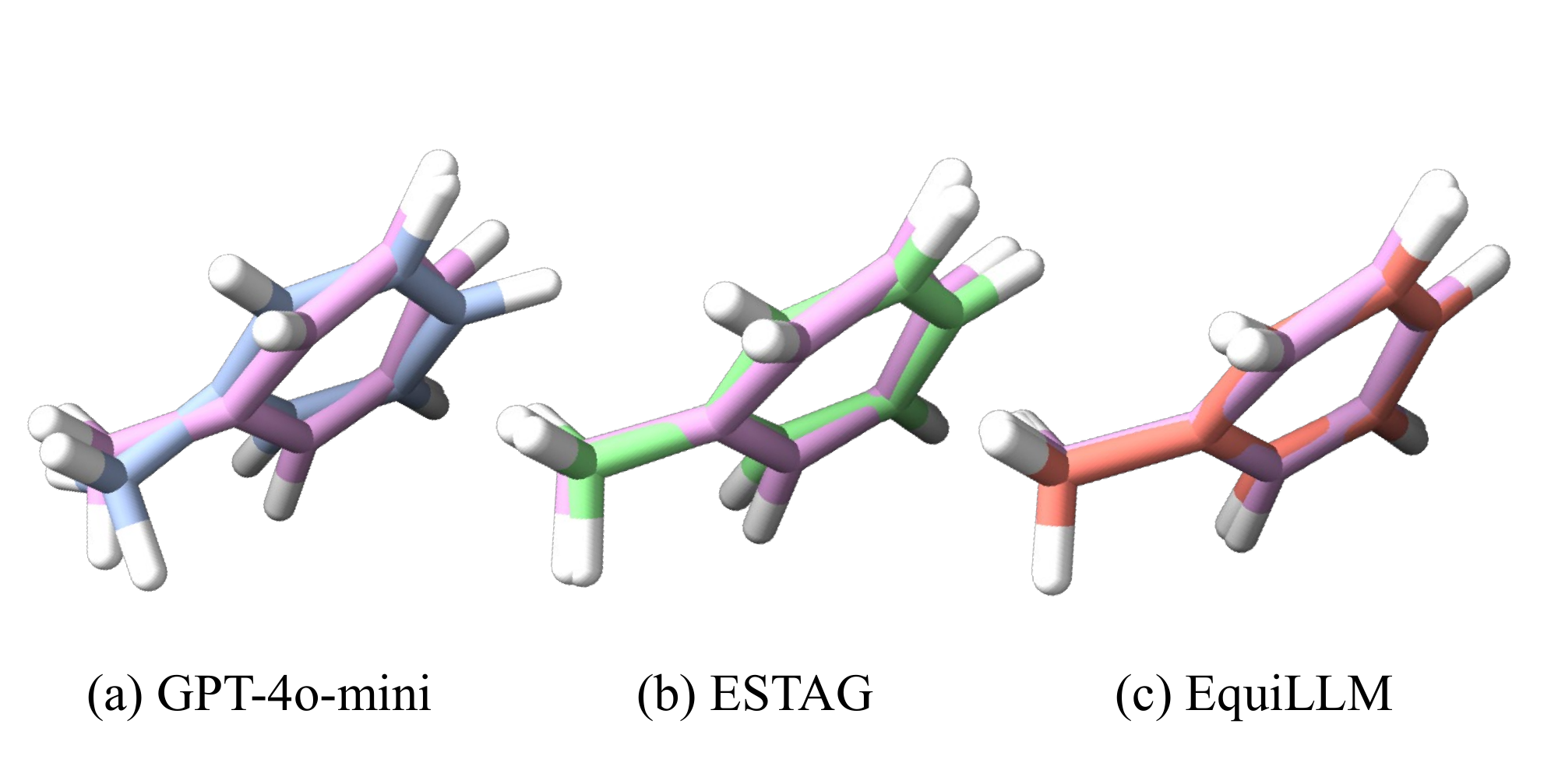} 
  \vspace{-1cm}
  \caption{The visualization of the predicted structures across various methods on Toluene of the MD17 dataset, where the pink represents the ground-truth structure.}
  \label{fig:rna_pdf}
\end{figure}

\textbf{Results.}
\cref{table:md17_r10} presents the performance of all models on MD17 dataset under the setting of predicting 10 frames from an input of 10 frames. From the table, the following conclusions can be drawn: \textbf{1.} The proposed EquiLLM framework achieves state-of-the-art (SOTA) performance on all eight molecules, demonstrating its superiority; \textbf{2.} Compared to our Equivariant Encoder model ESTAG, EquiLLM achieves a performance improvement of 5.41\% to 42.76\% on six molecules, indicating that EquiLLM effectively leverages knowledge from LLMs to enhance the prediction of molecular dynamics trajectories; \textbf{3.} We also tested the prediction capability of several leading LLMs, including GPT-4o-mini, Gemini-1.5-flash-latest, and DeepSeek-V3. DeepSeek-V3$^{\star}$ indicates that due to frequent API timeouts, several molecules do not complete the full testing dataset. Using the same prompt as EquiLLM, we provide the 3D coordinates of all atoms from the past 10 frames to these LLMs, allowing it to predict the coordinates of all atoms in the following 10 frames. The result shows that these LLMs significantly underperforms most baseline methods in prediction accuracy, indicating its weaker capability in directly predicting 3D coordinates. In contrast, our EquiLLM framework, by providing structured molecular descriptions and statistical constraints, enables the LLM to combine its pretrained knowledge with the specific task, thus significantly reduce the predicted MSE.

\subsubsection{Human Motion Simulation}
\begin{table*}[!t]
\begin{minipage}[c]{\textwidth}
\vspace{-0.5cm}
\begin{minipage}[c]{0.65\textwidth}
\begin{table}[H]
\centering
\caption{Predicted MSE ($\times 10^{-2}$) on Motion dataset.}
\label{table:motion}
\resizebox{\linewidth}{!}{
\begin{tabular}{lcccccc}
\toprule
\multirow{2}{*}{Method} & \multicolumn{3}{c}{Walk} & \multicolumn{3}{c}{Basketball} \\
\cmidrule(lr){2-4} \cmidrule(lr){5-7}
 & R=10 & R=15 & R=20 & R=10 & R=15 & R=20 \\
\midrule
\addlinespace[0.3em]
ST\_GNN & 1.150\tiny{$\pm$0.001} & 2.544\tiny{$\pm$0.814} & 2.765\tiny{$\pm$0.032} & 34.536\tiny{$\pm$0.747} & 133.731\tiny{$\pm$11.677} & 278.246\tiny{$\pm$2.225} \\
\addlinespace[0.3em]
ST\_TFN & 9.584\tiny{$\pm$0.156} & 20.667\tiny{$\pm$1.533} & 31.437\tiny{$\pm$0.120} & 168.674\tiny{$\pm$0.556} & 358.881\tiny{$\pm$1.661} & 613.755\tiny{$\pm$3.256} \\
\addlinespace[0.3em]
ST\_GCN & 18.737\tiny{$\pm$1.351}  & 19.467\tiny{$\pm$0.577} & 20.498\tiny{$\pm$2.232} & 275.744\tiny{$\pm$13.322} & 516.462\tiny{$\pm$91.545} & 662.488\tiny{$\pm$21.859} \\
\addlinespace[0.3em]
ST\_SE(3)-Tr. & 5.248\tiny{$\pm$0.132}  & 10.869\tiny{$\pm$0.596} & 20.999\tiny{$\pm$0.156} & 178.677\tiny{$\pm$4.022} & 390.518\tiny{$\pm$3.260} & 621.004\tiny{$\pm$12.186} \\
\addlinespace[0.3em]
ST\_EGNN & 2.867\tiny{$\pm$0.011} & 4.189\tiny{$\pm$0.172}  & 8.644\tiny{$\pm$1.620} & 30.813\tiny{$\pm$0.122} & 72.963\tiny{$\pm$1.295} & 152.551\tiny{$\pm$1.466} \\
\midrule
\addlinespace[0.3em]
ESTAG & 0.709\tiny{$\pm$0.052} & 1.877\tiny{$\pm$0.211} & 3.464\tiny{$\pm$1.127} & 10.507\tiny{$\pm$0.073} & 33.636\tiny{$\pm$0.425} & 76.548\tiny{$\pm$0.916} \\
\addlinespace[0.3em]
EquiLLM & \textbf{0.539}\tiny{$\pm$0.011} & \textbf{1.300}\tiny{$\pm$0.134} & \textbf{2.213}\tiny{$\pm$0.160} & \textbf{9.438}\tiny{$\pm$0.202} & \textbf{30.371}\tiny{$\pm$0.068} & \textbf{72.233}\tiny{$\pm$0.954} \\
\addlinespace[0.3em]
Reduction w.r.t ESTAG &\textcolor{red}{-23.97\%}&\textcolor{red}{-30.74\%} &\textcolor{red}{-36.11\%} &\textcolor{red}{-10.17\%} &\textcolor{red}{-9.70\%} &\textcolor{red}{-5.63\%} \\
\bottomrule
\end{tabular}
}
\end{table}
\end{minipage}
\begin{minipage}[c]{0.34\textwidth}
\begin{table}[H]
\centering
\caption{Results on RAbD benchmark.}
\vspace{3pt}
\label{table:rabd}
\resizebox{\linewidth}{!}{
\begin{tabular}{lccc}
\toprule
Method & AAR $\uparrow$ & TM-score $\uparrow$ & RMSD $\downarrow$ \\
\midrule
RosettaAD & 22.50\% & 0.9435  &  5.52 \\
\addlinespace[0.3em]
LSTM & 22.36\%  & -  &- \\
\addlinespace[0.3em]
C-LSTM  & 22.18\%  & -  & -  \\
\addlinespace[0.3em]
RefineGNN & 29.79\% & 0.8303 & 7.55 \\
\addlinespace[0.3em]
C-RefineGNN & 28.90\% & 0.8317 & 7.21 \\ 
\addlinespace[0.3em]
GeoAB  & 36.43\%  &\textbf{0.9836}& \underline{1.79}  \\
\midrule
\addlinespace[0.3em]
MEAN  & \underline{36.77\%} & 0.9812 & 1.81 \\
\addlinespace[0.3em]
EquiLLM  &\textbf{38.97 \%}  &\underline{0.9830}  & \textbf{1.73}  \\
\bottomrule
\end{tabular}
}
\end{table}
\end{minipage}
\end{minipage}
\end{table*}

\begin{table*}[!t]
\vspace{-0.5cm}
\centering
\caption{Ablation studies ($\times 10^{-3}$) on MD17 dataset.}
\label{table:md17_ablation}
\tabcolsep=0.16cm
\begin{tabular}{lcccccccc}
\toprule
 & Aspirin  & Benzene & Ethanol & Malonaldehyde & Naphthalene & Salicylic & Toluene & Uracil\\
\midrule
\addlinespace[0.3em]
EE (ESTAG) &3.263	&0.891	&1.090	&2.046 &2.036 &3.134	&1.634	&1.852\\
\addlinespace[0.3em]
LLM (w/o Prompt) + EE &2.524	&0.810	&1.062	&2.326	&2.363	&2.180	&1.538	&1.425\\
\addlinespace[0.3em]
LLM + EE &3.083	&0.773 &1.128	&2.100	&2.899	&2.581	&1.756	&1.418\\
\midrule
\addlinespace[0.3em]
w/o Prompt &3.671 &0.860	&1.092	&2.479	&2.837	&2.193	&1.941	&1.542\\
\addlinespace[0.3em]
w/o Object Feature &3.122	&0.833	&1.080	&2.255	&2.297	&2.470	&1.627	&1.387\\
\addlinespace[0.3em]
w/o Statistics & 3.532	&0.820	&1.054	&1.889	& 2.286 &2.528	&1.650 &1.463\\
\midrule
\addlinespace[0.3em]
EquiLLM &\textbf{2.391}	&\textbf{0.732}	&\textbf{1.031}	&\textbf{1.671}&\textbf{1.453}	&\textbf{2.162}&\textbf{1.178}	&\textbf{1.060}\\
\bottomrule
\end{tabular}
\vspace{-0.5cm}
\end{table*}

\textbf{Implementation details.}
The Human Motion Capture dataset contains human motion trajectory data across multiple scenes. We focus primarily on two sub-datasets: Subject \#35 (\texttt{Walk}) and Subject \#102 (\texttt{Basketball}). To ensure a fair comparison, all hyperparameters (\emph{e.g.} learning rate, number of training epochs) are kept consistent across our model and all other baselines. We utilize the ESTAG as the Equivariant Encoder and GPT-2 as the language model within our EquiLLM framework.

\textbf{Results.}
\cref{table:motion} presents a performance comparison of all models on the \texttt{Walk} and \texttt{Basketball} datasets under settings requiring the prediction of 10, 15, and 20 frames, respectively. From the table, it is evident that EquiLLM framework achieves SOTA performance across all six settings, with a performance improvement ranging from 5.63\% to 36.11\%. This demonstrates that EquiLLM effectively handles predictions over varying prediction lengths, exhibiting excellent robustness and generalization ability

\subsection{Antibody Design}
\label{sec:4.2}
Following previous study MEAN~\citep{kong2022conditional}, we selected complete antibody-antigen complexes from the SAbDab~\citep{dunbar2014sabdab} dataset to construct the training and validation sets. First, we performed clustering based on CDRs, grouping complexes with CDR sequence identity above 40\% into the same cluster. Then, the training and validation sets were partitioned in the same manner as in MEAN. For test set, we selected 60 diverse complexes from the RAbD~\citep{adolf2018rosettaantibodydesign} dataset to evaluate the performance of different methods. Before starting the experiments, we remove samples from the training and validation sets that belong to the same cluster as the test set to prevent data leakage.

\textbf{Baselines and metrics.}
We compared our EquiLLM with seven methods, including RosettaAD~\citep{adolf2018rosettaantibodydesign}, LSTM~\citep{saka2021antibody, akbar2022silico}, RefineGNN~\citep{jin2022iterative}, MEAN~\citep{kong2022conditional}, GeoAB~\citep{lin2024geoab}, and two variants of LSTM and RefineGNN, C-LSTM and C-RefineGNN, which utilize the full contextual information. We use AAR and RMSD to reflect the recovery ratio of the CDR-H3 amino acid sequence and the accuracy of the corresponding 3D structure prediction. Additionally, we employ the TM-score~\citep{zhang2004scoring, xu2010significant} to measure the global similarity between two protein structures. We utilize the MEAN~\citep{kong2022conditional} as the Equivariant Encoder and GPT-2 as the language model within our EquiLLM framework.

\textbf{Results.}
\cref{table:rabd} presents the performance of all models on the RAbD dataset. It can be concluded from the table: \textbf{1.} The proposed EquiLLM framework achieves the best performance in both AAR and RMSD metrics, with comparable results in the TM-score metric. The SOTA method GeoAB achieves superior performance in TM-score by incorporating more detailed geometric constraints, such as bond lengths, bond angles, and torsion angles, into the model, which enhances its overall structural prediction capabilities; \textbf{2.} Compared to the Equivariant Encoder model MEAN, EquiLLM shows significant improvement across all metrics, demonstrating that EquiLLM successfully leverages LLM’s knowledge integration and constraint-handling abilities while effectively utilizing LLM’s capacity for understanding and reasoning 1D sequences.

\subsection{Ablation studies}
\label{sec:4.3}
In this section, we delve into the design of the EquiLLM framework, analyzing the impact of different architectural designs and prompt configurations on model performance. The experimental results are shown in \cref{table:md17_ablation}, where EE represents the Equivariant Encoder.

\textbf{Architecture Design.} \textbf{1.} The results in the second row indicate that processing raw features through the LLM before feeding them into the Equivariant Encoder, while omitting the Equivariant Adapter for a simpler architecture, yields a performance improvement. This finding validates that the LLM has a fundamental capability to process and integrate structured information effectively. \textbf{2.} However, to fully exploit the potential of the LLM model, it is necessary to leverage prompts to capitalize on its strengths in text understanding. Building upon the second-row model, we perform experiments by adding prompts, as the results shown in the third row. The results indicate a significant performance drop. We speculate that this is due to the large semantic space difference between the unprocessed raw feature $\mH$ and the text features, which hampers the model's prediction capabilities. This suggests that LLM requires an appropriate interface to harness its advantages. This led to the design of the current EquiLLM framework.

\textbf{Prompt Design.} \textbf{3.} From \cref{table:md17_ablation}, it is evident that either completely removing the prompt or reducing its content (such as the Object Feature or Statistics) leads to a decline in performance. This observation reinforces our design philosophy: LLMs necessitate comprehensive information, including molecular descriptions and statistical constraints, to fully utilize their knowledge integration and constraint reasoning capabilities. This, in turn, facilitates more accurate predictive guidance.

\section{Conclusion}

We present EquiLLM, a framework that synergizes the strengths of LLMs and geometric GNNs to address the dual challenges of $\mathrm{E}(3)$-equivariance and knowledge integration in 3D physical system modeling. By introducing geometry-aware prompting and a modular architecture that isolates invariant and equivariant processing, EquiLLM circumvents the inherent limitations of LLMs in spatial reasoning while enabling the infusion of domain-specific knowledge through flexible prompting strategies. The separation of roles—LLMs as invariant feature processors and geometric GNNs as directional information handlers—provides a principled approach to preserving symmetry constraints. In future work, we plan to explore optimal prompting strategies for better leveraging domain knowledge and extending this framework to broader scientific tasks. We hope the EquiLLM framework will serve as a valuable reference for applying LLMs in scientific domains.

\bibliography{main}
\bibliographystyle{utils/icml2025}

\newpage
\appendix
\onecolumn

\end{document}